\documentclass[11pt,a4paper]{article}
\usepackage[hyperref]{acl2019}
\usepackage{times}
\usepackage{latexsym}

\usepackage{amsmath,amsfonts,bm}

\def\figref#1{figure~\ref{#1}}
\def\Figref#1{Figure~\ref{#1}}

\def\Secref#1{Section~\ref{#1}}

\def\eqref#1{equation~\ref{#1}}

\def\1{\bm{1}}

\def\ry{{\textnormal{y}}}

\def\va{{\bm{a}}}
\def\vb{{\bm{b}}}
\def\vc{{\bm{c}}}
\def\vd{{\bm{d}}}
\def\ve{{\bm{e}}}

\def\vh{{\bm{h}}}

\def\vs{{\bm{s}}}

\def\vx{{\bm{x}}}

\def\mA{{\bm{A}}}

\def\mE{{\bm{E}}}

\def\mH{{\bm{H}}}

\def\mM{{\bm{M}}}

\def\mW{{\bm{W}}}
\def\mX{{\bm{X}}}

\DeclareMathAlphabet{\mathsfit}{\encodingdefault}{\sfdefault}{m}{sl}
\SetMathAlphabet{\mathsfit}{bold}{\encodingdefault}{\sfdefault}{bx}{n}
\newcommand{\tens}[1]{\bm{\mathsfit{#1}}}

\def\tW{{\tens{W}}}

\newcommand{\softmax}{\mathrm{softmax}}

\newcommand{\sigmoid}{\sigma}

\DeclareMathOperator*{\argmax}{arg\,max}

\DeclareMathOperator{\real}{\rm I\!R}

\usepackage{xspace}

\newcommand{\red}[1]{\textcolor{red}{#1}}
\newcommand{\blue}[1]{\textcolor{blue}{#1}}

\newcommand{\green}[1]{\textcolor{green}{#1}}

\newcommand{\ie}{{\em i.e.,}\xspace}
\newcommand{\eg}{{\em e.g.,}\xspace}

\newcommand{\Ni}{({\em i})~}
\newcommand{\Nii}{({\em ii})~}

\newcommand{\Na}{({\em a})~}
\newcommand{\Nb}{({\em b})~}

\usepackage{url}

\usepackage{hyperref}
\usepackage{url}
\usepackage{graphicx}
\usepackage{amssymb}
\usepackage{comment}
\usepackage{enumitem}
\usepackage{caption}
\usepackage{subcaption}
\usepackage{booktabs}
\usepackage{tikz}
\usetikzlibrary{trees}

\usepackage[export]{adjustbox}

\aclfinalcopy 
\def\aclpaperid{1889} 

\title{A Unified Linear-Time Framework for Sentence-Level Discourse Parsing}

\author{Xiang Lin$^*$$^\P$, Shafiq Joty\thanks{*equal contribution} $ ^\P$$^\S$, Prathyusha Jwalapuram$^\P$ and M Saiful Bari$^\P$\\
  $^\P$Nanyang Technological University, Singapore \\
  $^\S$Salesforce Research Asia, Singapore \\
  \texttt{\{linx0057@e., srjoty@, jwal0001@e., bari0001@e.\}ntu.edu.sg } \\}
\date{}

\begin{document}
\maketitle

\begin{abstract}

We propose an efficient neural framework for sentence-level discourse analysis in accordance with Rhetorical Structure Theory (RST). Our framework comprises a discourse segmenter to identify the elementary discourse units (EDU) in a text, and a discourse parser that constructs a discourse tree in a top-down fashion. Both the segmenter and the parser are based on Pointer Networks and operate in linear time. Our segmenter yields an $F_1$ score of 95.4, and our parser achieves an $F_1$ score of 81.7 on the aggregated labeled (relation) metric, surpassing previous approaches by a good margin and approaching human agreement on both tasks (98.3 and 83.0 $F_1$).  


\end{abstract}

\newcommand{\fix}{\marginpar{FIX}}
\newcommand{\new}{\marginpar{NEW}}

\section{Introduction} \label{sec:intro}

Coherence analysis of a text is a fundamental task in Natural Language Processing that can benefit many downstream applications. Rhetorical Structure Theory or RST \cite{Mann88} is one of the most influential theories of text coherence. According to RST, a text is represented by a hierarchical structure known as a Discourse Tree (DT). As exemplified in Figure \ref{fig:example}, the leaves of a DT correspond to contiguous atomic text spans called Elementary Discourse Units (EDUs). The adjacent EDUs and larger units are recursively connected by certain coherence relations (\eg\ {\sc{Attribution}}, {\sc{Explanation}}). The discourse units connected by a relation are further categorized based on their relative importance: {\sc{Nucleus}} refers to the core part(s), while {\sc{Satellite}} refers to the peripheral one. Coherence analysis in RST involves two subtasks: \Ni breaking the text into a sequence of EDUs, referred to as \textbf{Discourse Segmentation}, and \Nii linking the EDUs into a DT, referred to as \textbf{Discourse Parsing}.     

\begin{figure}
\begin{tikzpicture}[
  tlabel/.style={pos=0.4,right=-1pt,font=\footnotesize\color{red!80!black}}, level 1/.style={sibling distance=35mm}, level 2/.style={sibling distance=30mm}
]
\node{\sc{\blue{Condition}}}
[level distance=10mm]
child {node {\sc{\blue{Attribution}}}
    child {node {$e_1$}
    edge from parent node[tlabel,left=2pt] {S}
    }
     child {node {$e_2$}
    edge from parent node[tlabel,left=2pt] {N}
    }
edge from parent node[tlabel,right=2pt] {N} 
}
child {node {\sc{\blue{Temporal}}}
    child {node {$e_3$}
    edge from parent node[tlabel,left=2pt] {N}
    }
     child {node {$e_4$}
    edge from parent node[tlabel,left=2pt] {N}
    }
edge from parent node[tlabel,right=2pt] {S} 
};


\end{tikzpicture}

\begin{tikzpicture}

\node[align=left] at (-1,-5) {\small{[The Treasury also said]$_{{e_1}}$ [noncompetitive tenders will be } \\ \small{considered timely]$_{e_2}$ [if postmarked no later than Sunday, }\\ \small{ Oct.29,]$_{e_3}$ [and received no later than tomorrow.]$_{e_4}$}};

\end{tikzpicture}
\vspace{-1.5em}
\caption{An example discourse tree with four EDUs.}
\label{fig:example}
\vspace{-1.2em}
\end{figure}

In this paper we consider {sentence-level} coherence analysis, which involves discourse segmentation and sentence-level parsing. For example, consider the DT in \figref{fig:example} for the sentence ``The Treasury also said noncompetitive tenders will be considered timely if postmarked no later than Sunday, Oct.29, and received no later than tomorrow.'', which has four EDUs as shown below the tree. Such sentence-level discourse annotations have been shown to be beneficial for a number of applications including machine translation  \cite{guzman-EtAl:ACL2014} and sentence compression \cite{Sporleder05}. Furthermore,  sentence-level analysis is considered to be a crucial step towards full text-level analysis. For example, automatic discourse segmentation has been shown to be the main source of inaccuracies in discourse parsing \cite{Marcu03,Joty-2012}, and sentence-level parsing is considered as an essential first step in many existing discourse parsers \cite{Feng-14-ACL,joty-carenini-ng-cl-15} including the state-of-the-art one \cite{Wang-acl-2017}. 


While earlier methods have mostly relied on hand-crafted lexical and syntactic features, recently researchers have shown competitive or even better results with neural models. One of the crucial advantages of \textbf{neural} models is that they can learn the feature representation of the discourse units in an end-to-end fashion. This capability is particularly enhanced through the use of effective pretrained word embeddings such as Glove \cite{pennington2014glove} that provide better generalization. Despite this, successful discourse parsers \cite{li-li-hovy:2014:EMNLP2014,ji-eisenstein:2014:P14-1,Li-16} still needed to use hand-engineered features to outperform the non-neural models.

Another important distinction between existing methods is whether they employ a greedy \textbf{transition-based} algorithm \cite{Marcu99,Feng12,Feng-14-ACL,ji-eisenstein:2014:P14-1,Braud-EACL17,Li-16, Wang-acl-2017} or a globally optimized \textbf{chart parsing} algorithm \cite{Marcu03,li-li-hovy:2014:EMNLP2014,joty-carenini-ng-cl-15}. Transition-based parsers build the tree incrementally by making a series of shift-reduce action decisions. The advantage of this method is that the parsing time is linear with respect to the number of EDUs \cite{Kenji-09}. The limitation, however, is that the decisions made at each step are based on local information, causing error propagation to subsequent steps. Also, when humans are asked to perform discourse analysis (segmentation and parsing), they tend to understand the full text first, before executing the tasks. 

Methods based on chart parsing, on the other hand, learn scoring functions for discourse subtrees and perform dynamic programming search over all  possible trees to find the most probable tree for a text. While these methods are more accurate than greedy parsers, they are generally slow, having a time complexity of $O(n^3 M)$  for $n$ EDUs and $M$ different relations \cite{joty-carenini-ng-cl-15}.  

In this paper, we propose a unified neural framework for discourse segmentation and parsing based on Pointer Networks \cite{Vinyals_NIPS2015}. Our parser employs a transition-based procedure to construct a discourse tree in a \textbf{top-down} fashion with the same computational efficiency, while still maintaining a global view of the input text. This is thanks to the \textbf{encoder-decoder} architecture that makes it possible to capture information from the whole text and the previously derived subtrees, while limiting the number of parsing steps to linear in the number of EDUs. Our framework is purely neural and does not rely on any hand-engineered features. Additionally, the framework allows us to  train the segmentation and parsing models seamlessly with a joint objective.

We conduct a series of experiments with our framework on the standard RST Discourse Treebank (RST-DT) dataset {for sentence level} parsing, and our main findings are as follows.

\begin{itemize}[leftmargin=*]

\item Our segmenter achieves an $F_1$ score of 95.4 giving a \emph{relative error reduction of  $32\%$} over the state-of-the-art segmenter. 

\item Evaluation of our  sentence-level discourse parser with manual segmentation shows that it achieves an $F_1$ score of 81.3 on the relation labeling task yielding a \emph{relative error reduction of about $17\%$} over the state-of-the-art parser. 
\item Joint training of the segmentation and parsing models improves the results further giving 95.5 $F_1$ on segmentation and 81.7 $F_1$ on parsing, while the human agreements on these two tasks are $98.3$ and $83$ $F_1$, respectively.

\item Our end-to-end system (segmenter$\rightarrow$parser) reaches an $F_1$ of 77.5 on relation labeling providing an \emph{absolute improvement of 10\%} compared to the best existing system. 

\item Both our discourse segmenter and parser operate in linear time with respect to the number of EDUs. In practice, our segmenter and parser individually give 6.79x and 3.92x speedups, while the end-to-end system gives 5.9x speedup compared to the best open-sourced system.

\end{itemize}

{We make our code available at \url{https://ntunlpsg.github.io/project/parser/pointer-net-parser}} 


\section{Background} 
\label{sec:background}
In this section, we give a brief overview of coherence analysis with RST and pointer networks. 

\subsection{Coherence Analysis with RST}

Coherence analysis has been a long standing problem. We give a brief overview of the studies that are directly related to our method. \citet{Marcu03} proposed {SPADE}, a system that uses generative models with syntactic features for discourse segmentation and sentence-level parsing. Subsequent research focuses on the impact of syntax in discourse analysis \cite{Sporleder05,Fisher07,Hernault10}. \citet{joty-carenini-ng-cl-15} propose CODRA, a system that comprises a discourse segmenter and a two-stage discourse parser -- one for sentence-level parsing and the other for multi-sentential parsing. \citet{Feng14_ACL} also propose two-stage parsing based on CRFs that use many hand-crafted features. 
\citet{li-li-hovy:2014:EMNLP2014} propose a recursive network for discourse parsing. \citet{ji-eisenstein:2014:P14-1} present a representation learning method in a shift-reduce discourse parser. \citet{Wang-acl-2017} propose a two-stage parser, where they use shift-reduce parsing to first construct a tree structure with only nuclearity labels, and then in the second stage they identify the relations. They use SVMs with a large number of features. \citet{wang2018edu} propose a discourse segmenter based on LSTM-CRF and achieve state-of-the-art results with ELMo. \citet{li-sun-joty-ijcai-18} also propose a segmenter based on pointer networks. 

Pointer networks have also been used for summarization \cite{See-17} and dependency parsing \cite{StackPTR}. In our work, we use pointer networks not only for segmentation but also for parsing, and we also show how the segmenter and parser can be trained jointly.

\subsection{Pointer Networks}

Sequence-to-sequence paradigms \cite{Sutskever_NIPS2014} provide the flexibility that the output sequence can be of a different length than the input sequence. However, they still require the output vocabulary size to be fixed a priori, which limits their applicability to problems where one needs to select (or point to) an element in the input sequence; that is, the size of the output vocabulary depends on the length of the input sequence. Pointer Networks \cite{Vinyals_NIPS2015} address this limitation by using \textbf{attention} \cite{BahdanauCB15} as a pointing mechanism. Specifically, an encoder network first converts the input sequence $\mX = (\vx_1, \ldots, \vx_n)$ into a sequence of hidden states $\mH = (\vh_1, \ldots, \vh_n)$. At each time step $t$, the decoder network receives the input from previous step and produces a decoder state $\vd_t$ that modulates an attention over inputs. The output of the attention is a softmax distribution over the inputs.  

\begin{eqnarray}
s_{t,i} = \sigmoid(\vd_t, \vh_i) \hspace{1em} \text{for } i = 1 \ldots n \\
\va_t = \softmax(\vs_t) = \frac{\exp(s_{t,i})}{\sum_{i=1}^n \exp(s_{t,i})} \label{eq:attn}
\end{eqnarray}
\normalsize

\noindent where $\sigmoid(.,.)$ is a scoring function for attention, which can be another neural network or simply a dot product. We use $\va_t$ for inferring the output: 
$\hat{\ry}_t = \argmax (\va_t) = \argmax p(\ry_t | \ry_{<t}, \mX, \theta)$, where $\theta$ is the set of model parameters. To condition on $\ry_{t-1}$, the corresponding input $\vx_{\ry_{t-1}}$ is copied as the input to the decoder. 



\section{Our Discourse Parser} \label{sec:par-model}

\begin{figure*}[t!]
    \centering
    \includegraphics[scale=0.30]{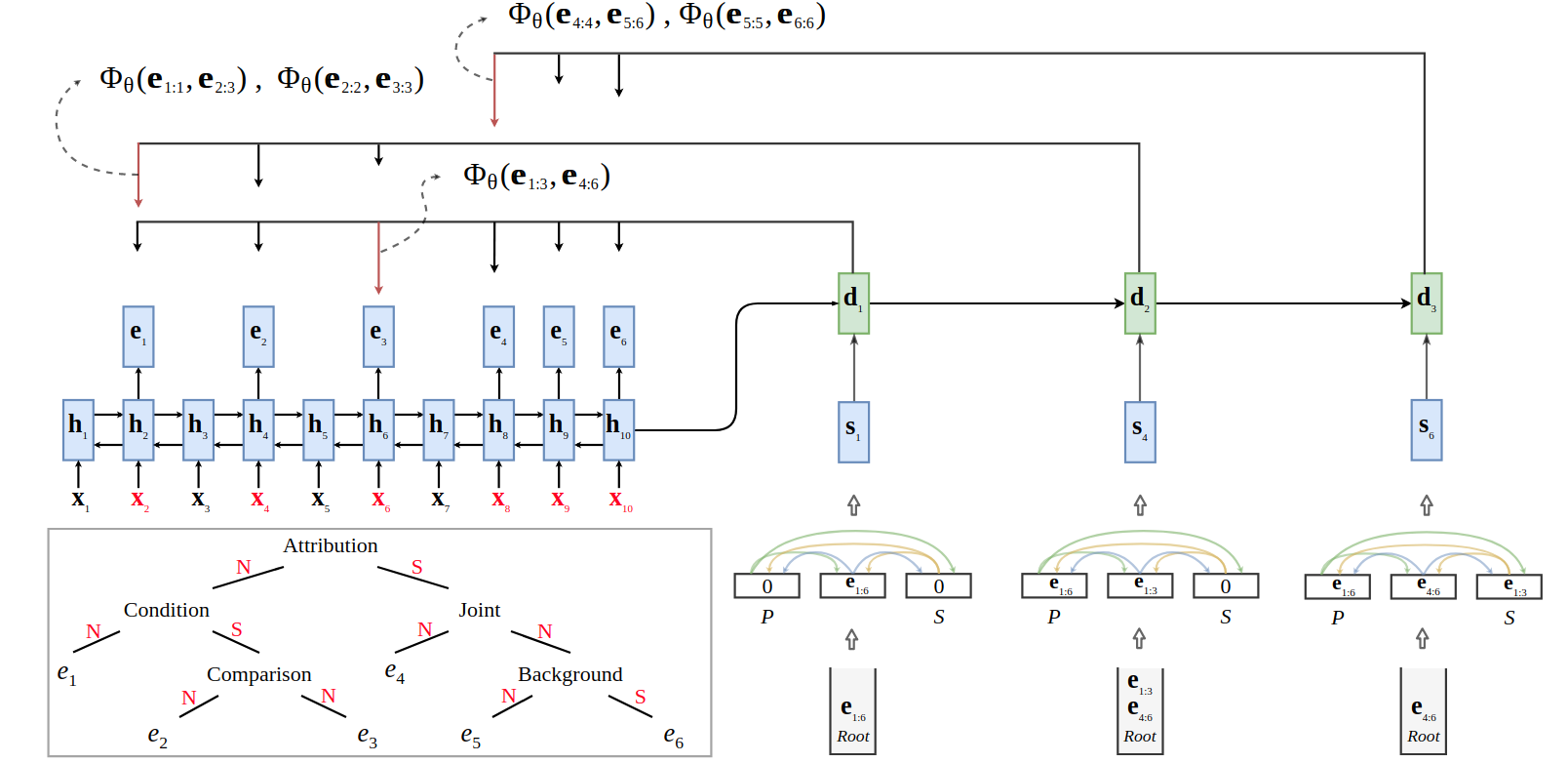}
    \caption{Our discourse parser along with the decoding process for a synthetic sentence with 10 words and 6 EDUs. EDU boundaries are marked in red color. For the inputs to the decoder at each step, $P$ and $S$ indicate the parent and sibling representations, respectively. $\Phi_\theta(\ve_{i:k}, \ve_{k+1:j})$ denotes the relation-nuclearity classifier employed by the parser to find the nuclearity and relation labels for the newly created spans, $\ve_{i:k}$ and $\ve_{k+1:j}$.}
    \label{fig:Parser}
\end{figure*}

Given a sentence as input, the framework first employs our discourse segmenter to break the sentence into a sequence of EDUs. Our parser then links these EDUs into a labeled tree by identifying \Ni which
discourse units to relate (\ie\ finding the right \textbf{structure} of the tree), and \Nii what relations and nuclearity statuses to use in connecting them (\ie\ finding the correct \textbf{labels}). In the interests of presentational simplicity, we first describe the discourse parser in this section assuming that the EDUs have already been identified.

\paragraph{Model Overview.}

As shown in Figure \ref{fig:Parser}, our parser uses a \textbf{pointer network} as its backbone parsing model. Given an input sentence containing $n$ words $(w_1, \ldots, w_n)$, we first embed the words into their respective distributed representation by initializing them either randomly or with pretrained embeddings such as Glove \cite{pennington2014glove} or ELMo \cite{Peters:2018}. 
The result of this is a sequence of word vectors $\mX = (\vx_1, \ldots, \vx_n)$, which is fed to the network. 


The encoder of the pointer network first composes the entire sentence sequentially into a sequence of hidden states $\mH = (\vh_1, \ldots, \vh_n)$. The last hidden state of each EDU (\eg\ $\vh_2, \vh_4,\vh_6, \vh_8, \vh_9$ and $\vh_{10}$ in Figure \ref{fig:Parser}) are selected to represent the corresponding EDU, thus, forming a sequence of EDU representations $\mE = (\ve_1, \ldots, \ve_m)$. From this, the  greedy decoder then constructs the discourse tree in a \textbf{top-down depth-first} manner.

The decoder maintains a \textbf{stack} $\mathcal{S}$ to keep track of the spans that need to be parsed further and their order (depth-first). $\mathcal{S}$ is initialized with the special \emph{Root} symbol. At each decoding step $t$, the decoder extracts a span $\ve_{i:j}$ from the top of $\mathcal{S}$, and uses the EDU representation $\ve_{j}$ to generate a decoder hidden state $\vd_t$, which is in turn used to compute the attention scores over the EDU representations in the selected range of spans ($\ve_{i}$ to $\ve_{j}$). Based on the attention scores, the decoder chooses a position $k$ in the range to generate a new split $(\ve_{i:k},\ve_{k+1:j})$. The parser then applies a \textbf{relation classifier} $\Phi_{\theta}(\ve_{i:k},\ve_{k+1:j})$, parameterized by $\theta$, on the new split  to predict the relation and the nuclearity labels. If the length of any of the newly created spans ($\ve_{i:k}$ and $\ve_{k+1:j}$) is larger than two, the parser pushes it onto the stack. For the span containing only two EDUs, the parser would automatically run the classifier $\Phi_{\theta}$ to predict the relation and nuclearity between the two EDUs.

Since the parser works in a depth-first manner, a text span is not parsed until a complete subtree for the preceding span is built (assuming we process the leftmost child first). This allows the decoder to exploit information from the generated subtrees in addition to the representation of the span being parsed. In the following paragraphs, we describe the components of our parser in detail. 



\paragraph{The Encoder.}

Our parser uses a recurrent neural network (RNN) based on bidirectional Gated Recurrent Units or BiGRU \cite{ChoGRU} as the encoder. Like LSTM \cite{hochreiter1997long}, GRU cells are also designed to capture long range dependencies, but have fewer parameters than LSTM cells. In particular, our encoder uses six (6) recurrent layers of BiGRU cells, and generates hidden states $\mH = (\vh_1, \ldots, \vh_n)$ by composing the word representations sequentially from left-to-right and from right-to-left, which is, $\vh_i = [\overset{\rightarrow}{\vh_{i}}; \overset{\leftarrow}{\vh_{i}}]$ with $\overset{\rightarrow}{\vh}_{i}$ and $\overset{\leftarrow}{\vh}_{i}$ being the forward and the backward states. The last hidden states of an EDU are used as the EDU representation, generating a sequence of EDU representations $\mE = (\ve_1, \ldots, \ve_m)$ for the input sentence.





\paragraph{The Decoder.}

Our parser uses a six-layer unidirectional GRU as the decoder. Instead of using the word embeddings, we feed our decoder with the corresponding encoder states for the span. This is because the encoder states contain more contextual information than the word embeddings \cite{StackPTR}. We use the representation of the last EDU  as the representation of the span. For example, span $\ve_{1:3}$ in Figure \ref{fig:Parser} is represented by $\ve_3$ (or $\vh_6$). We also experimented with taking the \emph{mean} of the corresponding hidden states (\eg\ $\text{mean}(\vh_1, \ldots, \vh_6)$ for $\ve_{1:3}$). We found the former to perform better in our experiments. 

At each decoding step $t$, the decoder combines the span representation with its previous state $\vd_{t-1}$ to generate the current state $\vd_t$, which is then used to compute the attentions over the corresponding encoder states ($\ve_{1}, \ldots, \ve_{3}$ for $\ve_{1:3}$). We use the simple \textbf{dot product} as the scoring function (\ie\ $\sigmoid(\vd_t, \vh_i)$ in Equation \ref{eq:attn}).    





\vspace{0.5em}
\noindent \textit{\textbf{Remark:}} In our earlier attempts, we experimented with a self-attention based encoder-decoder with positional encoding similar to \cite{vaswani2017attention} to reduce the encoding time from $O(n)$ (linear) to $O(1)$ (constant) time. However, the performance was inferior to the RNN-based encoder.




\subsection{The Relation Classifier}

For relation labeling, we adopt a \textbf{bi-affine} classifier. The classifier is a two-layer neural network that takes two spans $\ve_{i:k}$ and $\ve_{k+1:j}$ as input and predicts the corresponding relation label and the nuclearity statuses. {As before, we consider the representation of the last EDU as the representation of the span ($\ve_{k}$ for $\ve_{i:k}$ and $\ve_{j}$ for $\ve_{k+1:j}$).} The first layer is a dense layer with Exponential Linear Unit (ELU)  activations that maps the span representations $\ve_{k}$ and $\ve_{j}$ to latent label-specific features $\vc_{k}$ and $\vc_{j}$ of dimensions $d$. 

\begin{eqnarray}
\vc_k = \text{ELU}(\ve_k^T U_1);\hspace{1em} \vc_j = \text{ELU}(\ve_j^T U_2) \label{eq:dense}
\end{eqnarray}
\normalsize

The second layer is a bi-affine layer with a $\softmax$ activation to get a multinomial distribution over the relation labels:

\begin{multline}
P_{\theta_l}(\ry|\mX) = \softmax (\vc_k^T \tW_{kj} \vc_j + \vc_k^T\mW_k + \\
\vc_j^T \mW_j + \mathbf{b}) \label{eq:biaffine}   
\end{multline}
\normalsize

\noindent where $\tW_{kj} \in \real^{d \times d \times R}, \tW_{k} \in \real^{d \times R}$, and $\tW_{j} \in \real^{d \times R}$ are the weights and $\vb$ is a bias vector with $R$ being the number of relation labels. The bi-affine layer not only does a linear transformation of $\vc_k$ and $\vc_j$ but also models the correlation between $\vc_k$ and $\vc_j$ vectors \cite{DozatMann16}.  

Following previous work, we attach the nuclearity statuses with the relation labels. For example, in Figure \ref{fig:example}, the {\sc{Attribution}} relation between $\ve_1$ as a satellite and $\ve_2$ as a nucleus is jointly represented as {\sc{Attribution-SN}}. This representation allows us to perform the two tasks - relation identification and nuclearity assignment jointly.



\subsection{Incorporating Partial Tree Information}

As mentioned before, parsing a tree in a depth-first manner allows us to incorporate partial tree information while decoding a span. In this work, we consider information from the \textbf{parent} ($P$) and the immediate \textbf{left-sibling} ($S$) of the span being parsed ($E$). For example, in Figure \ref{fig:Parser}, when parsing span $\ve_{4:6}$, in addition to the current span, we consider its parent span $\ve_{1:6}$ (represented by $\ve_6$) and its left subtree span $\ve_{1:3}$ (represented by  $\ve_3$). As the relative importance of the three components may vary, we put a self-attention layer before feeding them to the decoder. Formally, we put them as rows in a matrix $\mM=[P;E;S]$ and perform:





\begin{eqnarray}
\mA = \softmax (\mM\mM^T)\mM \label{eq:selfatten}
\end{eqnarray}
\normalsize

{We take an element-wise sum of the three (row) vectors in $\mA$ to form the final span representation ($\vs$ in the figure) and feed it to the decoder.}

\subsection{Training Loss}

Our parser is trained to minimize the sum of the loss for building the right tree \textbf{structure} and the loss for finding the correct \textbf{labels}. The \emph{structure loss} $\mathcal{L}_s$ is the pointing loss for the pointer network:

\begin{eqnarray}
\mathcal{L}_s(\theta_s) = 
\displaystyle - \sum_{t=1}^{T} \log P_{\theta_s}(\ry_t | \ry_{<t}, \mX)  \label{eq:LossParser}
\end{eqnarray}
\normalsize

\noindent where $\theta_s$ denotes the parameters of the encoder and the decoder, $y_{<t}$ represents the subtrees that have been generated by our parser at previous steps, and $T$ is the number of spans containing more than two EDUs (pushed in the stack).

The \emph{label loss} $\mathcal{L}_l$ is the cross entropy loss for the relation classifier, and can be defined as:



\begin{eqnarray}
\mathcal{L}_l(\theta_l) = 
\displaystyle - \sum_{i=1}^{I} \sum_{r=1}^{R} y_{i,r} \log P_{\theta_l}(\ry_i = r | \mX) \label{eq:LossClass}
\end{eqnarray}
\normalsize

\noindent where $\theta_l$ are the parameters for the relation classifier (including the encoder), $I$ is the number of spans with at least two EDUs, $R$ is the total number of relation labels, and $y_{i,r}$ is the one-hot encoding of the relation label. We also apply an $L_2$-regularization on the parameters. Hence, the final parsing loss  $\mathcal{L}_P(\theta_P)$ can be written as:


\begin{eqnarray}
\mathcal{L}_{\text{PAR}}(\theta_P) =  \mathcal{L}_s(\theta_s) + \mathcal{L}_l(\theta_l) + \frac{\lambda}{2} \| \theta_P\|^2_2 \label{eq:LossAll}
\end{eqnarray}
\normalsize

\noindent where $\lambda$ is the regularization strength and $\theta_P$ denotes the set of all parameters of the parser.



\section{Our Discourse Segmenter} \label{sec:seg-model}





{The job of the discourse segmenter is to find the EDU boundaries (if any) inside a sentence. } Traditionally, discourse segmentation has been treated either as an binary classification problem \cite{Marcu03,Fisher07} or as a sequence labeling problem \cite{wang2018edu}. Recently, \citet{li-sun-joty-ijcai-18} show the benefits of using pointer networks over previous methods for this task, achieving state-of-the-art results. In our work, we adopt their approach and advance the state-of-the-art further by simple modifications. More importantly, this framework allows us to train the discourse segmenter and the parser jointly with a shared encoder. 



\paragraph{Model Description.} 

\begin{figure}[t!]
    \centering
    \includegraphics[scale=0.1]{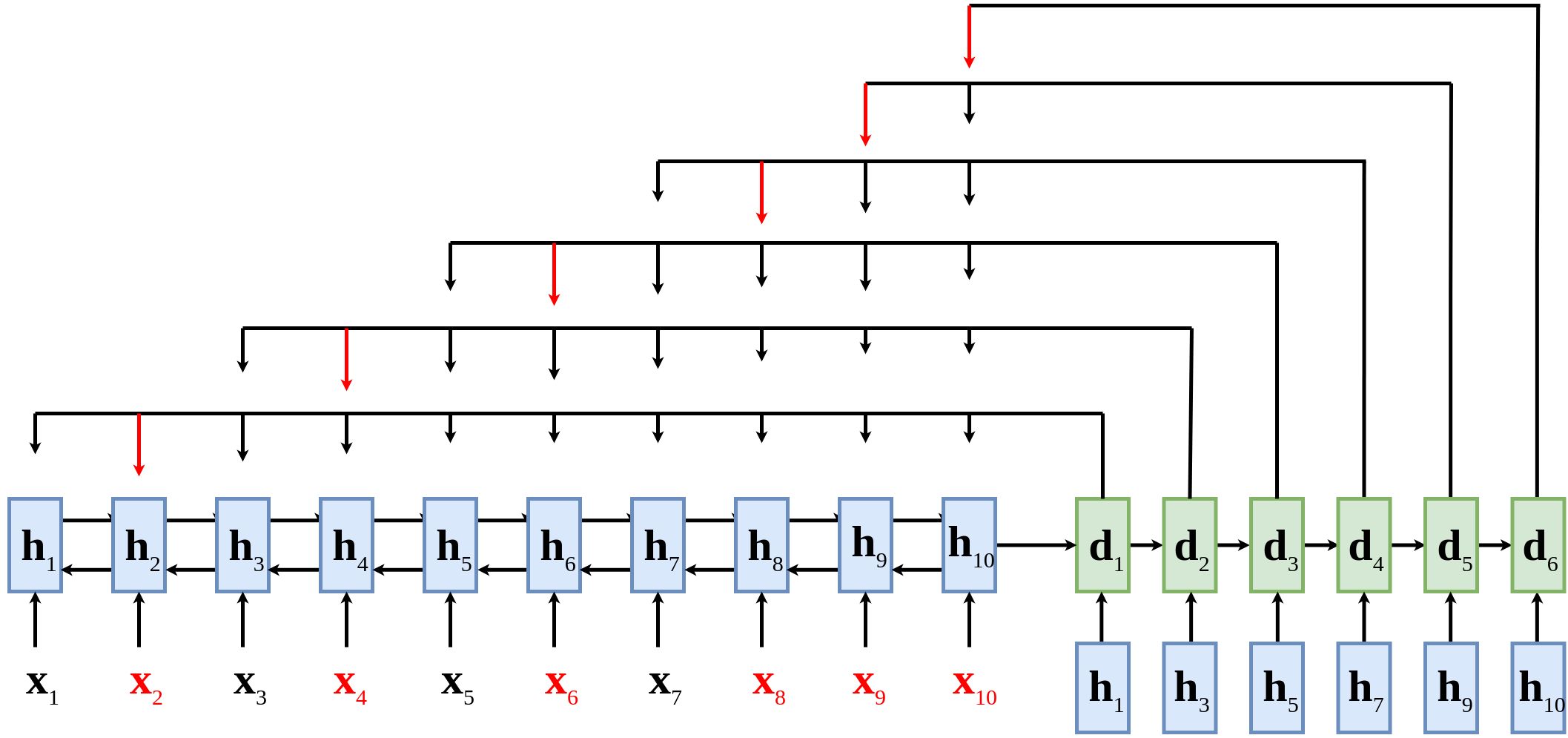}
    \caption{Our neural discourse segmentation model for the same synthetic sentence as in \Figref{fig:Parser}. Words in red color denote boundary words.} 
    \label{fig:Segmenter}
\end{figure}

\Figref{fig:Segmenter} depicts the architecture of our segmenter. Similar to our parser (\Figref{fig:Parser}), the encoder of our segmentation model reads the whole sentence and transforms it into a sequence of hidden states. Then, at each time step, the decoder receives an encoder state corresponding to the first token of a segment currently being processed, and produces a decoder state which is in turn used to compute a  distribution (attention) over all valid positions of the input sentence.  

The encoder and the decoder have the same architecture as in \cite{li-sun-joty-ijcai-18} with the following key improvements. First, following the same idea as in our parser, the decoder takes the encoder states as the input instead of word embeddings. Second, similar to our parser, we adopt dot product attention instead of an additive attention. Dot product attention is simple yet powerful, while using fewer parameters \cite{vaswani2017attention}. Third, instead of simple look-up based embedding methods such as Glove, we use the contextual embedding ELMo that captures rich contextual information.

We train the model by minimizing the pointing loss with an $L_2$-regularization on the weights.

\begin{equation}
\mathcal{L}_{\text{SEG}}(\theta_{S}) = 
\displaystyle - \sum_{j=1}^{J} \log P_{\theta_{S}}(\ry_j | \ry_{<j},\mX)  + \frac{\lambda}{2} \| \theta_{S}\|^2_2 \label{eq:LossSeg}
\end{equation}
\normalsize

\noindent where $\theta_{S}$ represents the model parameters and $J$ is the number of EDUs in a sentence.

\subsection{Joint and End-to-End Training}

One crucial advantage of our framework is that it allows us to train the segmentation and the parsing models simultaneously and/or in an end-to-end fashion, while sharing a common encoder. Intuitively, both discourse segmentation and parsing can benefit from each other -- a plausible segmentation can result in a plausible parse and vice versa. Such multitask learning was not possible in a non-neural setup and the two discourse analysis tasks have always been considered independently.


\begin{figure}
\centering
    \includegraphics[scale=0.16]{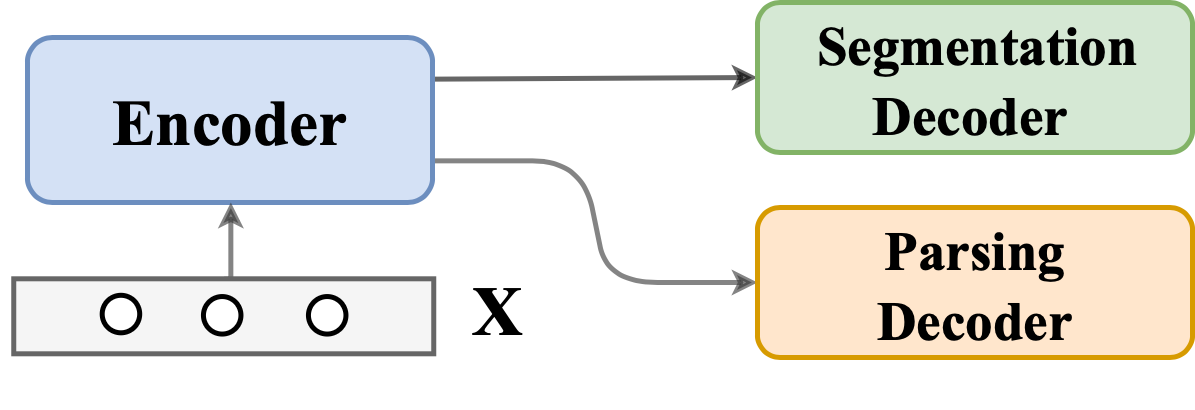}
    \caption{Joint training for segmentation and parsing. }
    \label{fig:joint}
\end{figure}

\Figref{fig:joint} depicts the schematic diagram of our joint training process. The segmentation and the parsing models share a common encoder while having two separate decoders for the two tasks. The training objective can be written as: 







\begin{eqnarray}
\mathcal{L}(\theta) =  \mathcal{L}_{\text{SEG}}(\theta) + \mathcal{L}_s(\theta) + \mathcal{L}_l(\theta) + \frac{\lambda}{2} \| \theta\|^2_2 \label{eq:LossJoint}
\end{eqnarray}
\normalsize

\noindent where $\theta$ denotes the parameters of our joint model. 


\section{Experiments} \label{sec:experiments}


In this section, we present our experiments on discourse segmentation and parsing.

\subsection{Datasets}

We train and evaluate our models on the standard RST Discourse Treebank (RST-DT) corpus \cite{Carlson02}. RST-DT contains discourse annotations for 385 news articles  from Penn Treebank \cite{Marcus93}. The training data  contains 347 documents (7673 sentences) and the test data contains 38 documents (991 sentences). In addition, 53 documents (1208 sentences)  were annotated by two human annotators, which we use to compute human agreement scores.

Since we focus on sentence-level discourse analysis, we follow the same setup as \citet{Marcu03,Joty-2012}. For segmentation, we utilize all 7673 sentences for training and 991 sentences for testing. For parsing, we extract sentence-level DTs from a document-level DT by finding the subtrees that span over the respective sentences. This gives 7321 sentence-level DTs for training, 951 for testing, and 1114 for getting human agreements. These numbers match the numbers reported by \citet{Joty-2012}. We randomly selected 10\% of the data from the training set for hyperparameter tuning.

\subsection{Discourse Segmentation Experiments}

\paragraph{Settings.} We compare our segmenter with five baselines: SPADE segmenter \cite{Marcu03}, F\&R \cite{Fisher07}, JCN \cite{Joty-2012}, SegBot \cite{li-sun-joty-ijcai-18}, and WLY \cite{wang2018edu}. 
Following the standard, we measure {$F_1$-score} based on the segmenter's ability to find the intra-sentential segment boundaries. 

When we evaluate the WLY segmenter on the standard test set using their released pretrained model,\footnote{\href{https://github.com/PKU-TANGENT/NeuralEDUSeg}{https://github.com/PKU-TANGENT/NeuralEDUSeg}} we get much lower results (90.5 $F_1$) than what they report in their paper (94.3 $F_1$). Upon investigation, we found that their experimental setting does not match with the standard one. Particularly, when extracting the sentences from the RST-DT dataset, instead of using gold tokenization, they use an automatic tokenizer, which gives fewer sentences -- 865 test sentences instead of 991 and 6132 training sentences instead of 7673. {This makes the scores artificially high.}\footnote{We  confirmed this by communicating with the authors.} 



For a fair comparison with our model, we train and evaluate WLY and SegBot on the same dataset setting, and report the mean and standard deviation of five runs, each run with a different random seed. WLY uses ELMo embeddings, which we also use in our model. 
To train our model, we use Adam optimizer with a batch size of 80. 
We apply $0.2$ dropout rate to the encoder and the decoder. The hidden sizes of the encoder, the decoder and the classifier are all set to $64$. {See Appendix for a complete list of hyperparameter settings.} 
In all our experiments when comparing two systems, we use the \emph{paired t-test} to measure statistical significance.


\renewcommand{\arraystretch}{1.1}
\begin{table}[t!]
\scalebox{0.62}{\begin{tabular}{l|cccc} 
\textbf{Approach} & \multicolumn{1}{c}{\bf{Precision}} & \bf{Recall} & \multicolumn{1}{c}{$\mathbf{F_1}$}\\
\midrule
\bf{Human Agreement} & 98.5 & 98.2 & 98.3 \\ 
\midrule
\bf{Baselines} & & & \\
SPADE \cite{Marcu03} &  83.8 &  86.8 &  85.2 \\
F\&R \cite{Fisher07} & 91.3 & 89.7 & 90.5 \\
JCN \cite{Joty-2012} & 88.0 & 92.3 & 90.1 \\
SegBot$_{\text{glove}}$ \cite{li-sun-joty-ijcai-18} 
& 91.08$_{\pm0.46}$  & 91.03$_{\pm0.42}$  & 91.05$_{\pm0.11}$  \\ 
WLY$_{\text{ELMo}}$ \cite{wang2018edu} & 92.04$_{\pm0.43}$ & 94.41$_{\pm0.53}$ & 93.21$_{\pm0.33}$\\
\midrule
\bf{Our Segmenter} & & & \\
Pointer Net (Glove) & 90.55$_{\pm0.33}$ & 92.29$_{\pm0.09}$ & 91.41$_{\pm0.21}$ \\
Pointer Net (BERT)  & 92.05$_{\pm0.44}$ & 95.03$_{\pm0.28}$ & 93.51$_{\pm0.16}$ \\
Pointer Net (ELMo) & {94.12$_{\pm0.20}$$^\star$} & {96.63$_{\pm0.12}$$^\star$} & {95.35$_{\pm0.10}$$^\star$} \\
~~+ Joint training & \textbf{93.34}$_{\pm0.23}$$^\star$ & \textbf{97.88}$_{\pm0.16}$$^\star$ &  \textbf{95.55}$_{\pm0.13}$$^\star$ \\ 
\bottomrule
\end{tabular}
}
\caption{Discourse segmentation results. Superscript $^\star$ indicates the model is significantly superior to the WLY$_{\text{ELMo}}$ model with a p-value $<0.01$.}
\label{table:seg-results}
\end{table}

\paragraph{Results.} Table \ref{table:seg-results} shows our segmentation results. As mentioned in  \Secref{sec:seg-model}, we implemented three key improvements on the top of \cite{li-sun-joty-ijcai-18}. Using encoder hidden states as decoder inputs and adopting dot product as the attention score function together gives  0.40\%-7.29\% relative improvement in $F_1$ over the first four baselines. Using ELMo, our segmenter outperforms all the baselines in all three measures. We achieve 2.3\%-11.9\%, 2.4\%-11.3\% and 2.3\%-12.3\% relative improvements in $F_1$, Recall and Precision, respectively. {Jointly training with the parser improves this further (95.55 $F_1$).} 
It is worthwhile to mention that our segmenter's performance of 95.55 $F_1$ is very close to the human agreement of 98.3 $F_1$. ELMo, as a transfer learning method, provides notable improvements. A similar observation was reported in \cite{wang2018edu}. Surprisingly, the results with BERT were not as good. We suspect this is due to BERT's special tokenization.





\subsection{Discourse Parsing Experiments}

\paragraph{Settings.}

We evaluate our parser in two different settings: 
\Na \textbf{parsing with gold segmentation}, and \Nb \textbf{parsing with our automatic segmentation} or \textbf{end-to-end evaluation}. In the first setting, we compare our results with SPADE \cite{Marcu03}, DCRF \cite{Joty-2012}, DPLP \cite{ji-eisenstein:2014:P14-1}, and the most recent 2-Stage Parser \cite{Wang-acl-2017}. SPADE and DCRF are both sentence-level parsers. 
However, DPLP and 2-Stage Parser are document-level parsers, and they do not report sentence-level performance. For DPLP, we feed the parser one sentence at a time to get a sentence-level DT. The 2-Stage Parser constructs a tree in multiple stages -- first sentence-level, then paragraph-level, and finally document-level. We ran their parser to generate all the document-level DTs in the test set, from which we extract the sentence-level DTs to evaluate. By our count, this gives 881 valid sentence-level trees as opposed to 951. This is because like their discourse segmenter (WLY), they use an  automatic tokenizer instead of gold tokenization. We evaluate their parser based on these 881 sentences. This is also what the authors suggested when contacted. 

In our second setting for full system evaluation, we compare with the two existing end-to-end systems, SPADE and DCRF. The hyperparameters (learning rate, batch size,  layer size) of our models in these two settings remain almost the same as the segmentation model (see Appendix for details).       


\paragraph{Metric and Relation Labels.}
We evaluate the performance by using the standard unlabeled (Span) and labeled (Nuclearity, Relation) {micro-averaged} precision, recall and $F_1$-score as described in \cite{Marcu00}. For brevity, we report only the $F_1$-scores here. Following previous work, we use the same 18 relations as used by previous studies, and we also attach the nuclearity statuses (NS, SN, NN) to these relations, giving a total of 39 distinctive relation labels.


\begin{table}[t!]
\scalebox{0.68}{\begin{tabular}{l|ccc}  
\textbf{Approach} & \multicolumn{1}{c}{\bf{Span}} & \bf{Nuclearity} & \bf{Relation}\\
\midrule
\bf{Human Agreement} & 95.7 & 90.4 & 83.0 \\ 
\midrule
\bf{Baselines} & & & \\
SPADE \cite{Marcu03} &  93.5 &  85.8 &  67.6 \\
DCRF \citep{Joty-2012} & 94.6  & 86.9  & 77.1  \\ 
DPLP \citep{ji-eisenstein:2014:P14-1} & 93.5  & 81.3  & 70.5 \\ 
2-Stage Parser \citep{Wang-acl-2017} & 95.6 & 87.8 & 77.6 \\
\midrule
\bf{Our Parser} & & & \\
Stack Pointer (ELMo-medium) & 96.37 & 89.04$^\star$ & 79.03$^\star$ \\
Stack Pointer (ELMo-large) & 96.86$^\star$ & 90.77$^\star$ & 81.12$^\star$ \\

~~+ Partial tree information & 96.94$^\star$ & 90.89$^\star$ & 81.28$^\star$ \\
~~+ Joint training & \textbf{97.44}$^\star$ & \textbf{91.34}$^\star$ &  \textbf{81.70}$^\star$ \\ 

\bottomrule
\end{tabular}
}
\caption{Parsing results with gold segmentation. Superscript $^\star$ indicates the model is significantly superior to the 2-Stage Parser with a p-value $<0.01$.}
\label{table:pargold-results}
\end{table}

\paragraph{Results with Gold Segmentation.}

We present the results in Table \ref{table:pargold-results}. Our base model (with ELMo-medium) outperforms all the existing methods to date in all three tasks. We achieve an absolute 1.43 $F_1$ improvement on the most difficult task of relation labeling, compared to the 2-stage parser (SOTA). Notably, the $F_1$ score of $96.37$ for Span of our base model even exceeds the human agreement ($F_1$ score of 95.7) on the doubly-annotated data. As it ought to be, incorporating full-size ELMo boosts the performance in three tasks.

\begin{figure}[t!]
\centering 
    \includegraphics[scale=0.24]{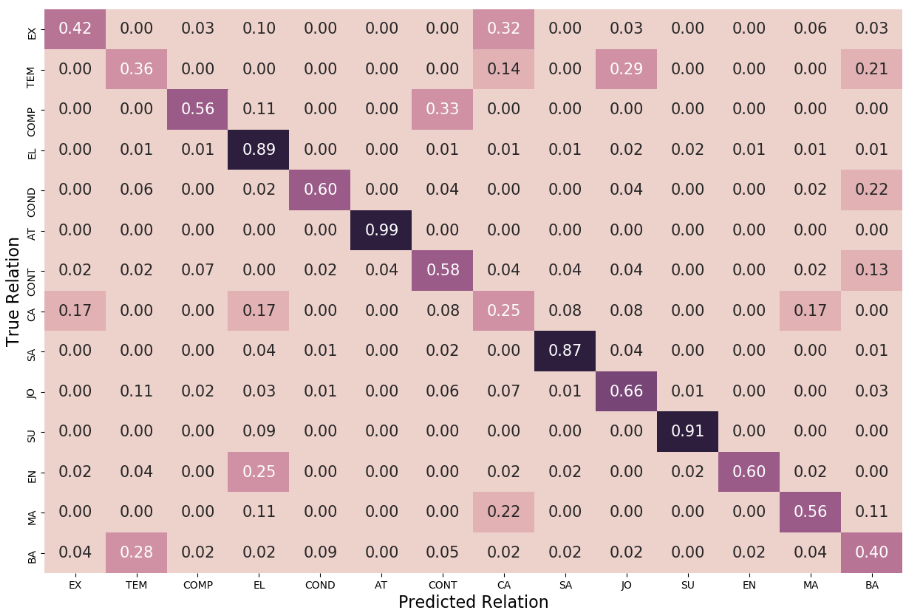}
    \vspace{-1em}
    \caption{Confusion matrix for 14 most frequent relation labels: \textbf{EX}PLANATION, \textbf{TEM}PORAL, \textbf{COMP}ARISON, \textbf{EL}ABORATION, \textbf{COND}ITION, \textbf{AT}TRIBUTION, \textbf{CONT}RAST, \textbf{CA}USE, \textbf{SA}ME-UNIT, \textbf{JO}INT, \textbf{SU}MMARY, \textbf{EN}ABLEMENT, \textbf{MA}NNER-MEANS, and \textbf{BA}CKGROUND.
    }
    \label{fig:confu}
\end{figure}


Our parser yields further improvements (+0.16 $F_1$ in Relation) by exploiting \textbf{partial tree} information generated in previous steps. The key component contributing to this improvement is the self-attention over original decoder inputs with partial tree information as described in Section \ref{sec:par-model}.\footnote{Simple averaging of the vectors did not show any gain.}   


Thanks to the pointer network as the backbone of our model, we are able to train our segmenter and parser jointly by sharing the same encoder. The last row of Table \ref{table:pargold-results} shows the results when we train the model jointly, and feed the parser with gold EDU segmentation during inference. The performance is improved further with \textbf{joint training}, achieving 97.44, 91.34, 81.70 $F_1$ score, in Span, Nuclearity and Relation, respectively. The results accord with our assumption that discourse segmentation and parsing may benefit from each other. Our parser surpasses human agreement in span and nuclearity. We are also approaching human agreement in the most difficult task of relation labeling. We show a confusion matrix for the relation labels in Figure \ref{fig:confu}. We see that our model gets confused between relations that are semantically similar (\eg\ \textbf{CA}USE vs. \textbf{EX}PLANATION,  \textbf{COMP}ARISON vs. \textbf{CONT}RAST, and \textbf{TEM}PORAL vs. \textbf{JO}INT). 


\vspace{0.5em}
\noindent \textit{\textbf{Remark:}} We observe that the relation labels in RST-DT are highly imbalanced, which makes the task harder. 
Therefore, we experimented with a variant of our parser where we had a separate classifier for nuclearity prediction, leaving 18 labels for relation classifier instead of 39. 
This model gave 96.74, 90.38, and 80.89 $F_1$ in Span, Nuclearity and Relation, respectively, which are lower than what we get by having a single classifier. Jointly modeling nuclearity and relation enforces the constraint that certain relations can have certain nuclearity orientations. For example, {\emph{Elaboration}} and {\emph{Attribution}} are mono-nuclear (takes either NS or SN), and Same–Unit and Joint are multi-nuclear relations (takes only NN).


\begin{table}[t!]
\centering
\scalebox{0.68}{\begin{tabular}{l|ccc}  
\textbf{Approach} & \multicolumn{1}{c}{\bf{Span}} & \bf{Nuclearity} & \bf{Relation}\\
\midrule
\bf{Baselines} & & & \\
SPADE \cite{Marcu03} & 76.7 & 70.2 & 58.0 \\
DCRF \citep{Joty-2012} & 82.4 & 76.6 & 67.5 \\ 
\midrule
\bf{Our Model} & & & \\
Stack Pointer (Pipeline) & 91.14$^\star$ & 85.80$^\star$ & 76.94$^\star$ \\
Stack Pointer (Joint training) & \textbf{91.75}$^\star$ & \textbf{86.38}$^\star$  & \textbf{77.52}$^\star$ \\
\bottomrule
\end{tabular}
}
\caption{Parsing results with automatic segmentation. Superscript $^\star$ indicates the model is significantly superior to the DCRF model with a p-value $<0.01$.}
\label{table:parseg-results}
\end{table}




\paragraph{End-to-End Performance.}


Table \ref{table:parseg-results} shows the results of our model and the two baselines. First, we use our segmenter followed by our best parser (independently trained) in a \textbf{pipeline}. The performance of this system is significantly better compared to the baselines. Against the best baseline (DCRF), it yields \textbf{8.74\%, 9.2\%, 9.44\%} absolute improvements in Span, Nuclearity, Relation, respectively. We push the performance even further by \textbf{joint training} of the segmenter and parser as in Figure \ref{fig:joint}. Notice that the performance on Relation (77.5 $F_1$) is even better than the DCRF model with gold segmentation (77.1 $F_1$) in Table \ref{table:pargold-results}.

\subsection{Run Time Analysis}

As noted earlier, both our segmenter and parser operate in linear time with respect to the number of input units. We compare the speed (sentences per second) of our systems against other baselines in Table \ref{table:speed-results} from a practical viewpoint. We test all the systems with the same 100 sentences randomly selected from our test set on our machine (CPU: Intel Xeon W-2133, GPU: NVIDIA GTX 1080Ti). We include the model loading time for all the systems.\footnote{As a neural model, WLY should be faster than the number we report. We retest both WLY and our model by excluding the model loading time. The speed of WLY and our segmenter are 157.80 sents/s and 181.30 sents/s, respectively. This could be because the two models are implemented in different frameworks (WLY: TensorFlow, ours: PyTorch).} Since SPADE and CODRA need to extract a handful of features, they are typically slower than the neural models which use pretrained embeddings. In addition, CODRA's DCRF parser has a $O(n^3)$  inference time. 
Our segmenter is 6.8x faster than SPADE. Compared to CODRA (the fastest parser as of yet), our parser is 3.9x faster. Finally, our end-to-end system is 5.9x faster than the fastest system out there (SPADE), making our system not only effective but also highly efficient. Even when tested only on CPU, our model is faster than all the other models.

\begin{table}[t!]
\centering
\scalebox{0.71}{\begin{tabular}{l|cc}  
\toprule
\textbf{System} & \bf{Speed (Sents/s)} & \bf{Speedup}\\
\midrule
\bf{Only Segmenter} & & \\
CODRA \cite{joty-carenini-ng-cl-15} & 3.06 & 1.0x \\ 
WLY \cite{wang2018edu} & 4.30 & 1.4x \\
SPADE \cite{Marcu03} & 5.24 & 1.7x\\
Our (CPU) & 12.05 & 3.9x \\
Our (GPU) & 35.54 & 11.6x \\
\midrule
\bf{Only Parser} & &  \\
SPADE \cite{Marcu03} & 5.07 & 1.0x  \\ 
CODRA \cite{joty-carenini-ng-cl-15} & 7.77 & 1.5x\\
Our (CPU) & 12.57 & 2.5x \\
Our (GPU) & 30.45 & 6.0x \\
\midrule
\bf{End-to-End (Segmenter $\rightarrow$ Parser)} & & \\
CODRA \cite{joty-carenini-ng-cl-15} & 3.05 & 1.0x \\ 
SPADE \cite{Marcu03} & 4.90 & 1.6x \\
Our (CPU) & 11.99 & 3.9x \\
Our (GPU) & 28.96 & 9.5x \\
\bottomrule
\end{tabular}
}
\caption{Speed comparison of our systems with other open-sourced systems.}
\label{table:speed-results}
\end{table}

\section{Conclusions}
\label{sec:conclusion}

We have proposed a unified framework for sentence-level discourse analysis based on pointer networks that constructs a discourse tree in linear time. Both our segmenter and parser achieve state-of-the-art results outperforming existing systems by a wide margin, without using any hand-crafted features. We also train the segmenter and the parser jointly through the encoder-decoder architecture and improve the results further. Apart from the effectiveness, our system is 6 times faster than the fastest available system.

Based on what we have done so far, it is natural for us to move our focus from sentence-level to document-level RST parsing. Also, in the future, we would like to investigate how our approach can be generalized to other discourse frameworks such as the Penn Discourse Treebank (PDTB).  





\section*{Acknowledgments}

We would like to extend our thanks for the funding support from MOE Tier-1 (Grant M4011897.020).

\bibliographystyle{acl_natbib}
\bibliography{refs}


\end{document}